%% file: main.tex
\documentclass[runningheads]{llncs}

\usepackage[mobile]{eccv}

\usepackage[table]{xcolor}
\usepackage{multirow,tabularx}
\usepackage{nicematrix}
\usepackage[usestackEOL]{stackengine}
\usepackage{xpatch}
\usepackage{comment}

\usepackage{eccvabbrv}
\usepackage{graphicx}
\usepackage{booktabs}
\usepackage{adjustbox}
\usepackage{pifont} 
\usepackage[font=small,skip=3pt,belowskip=-7pt]{caption}

\usepackage[accsupp]{axessibility}  %

\newcommand{\method}{AvatarPopUp\xspace}

\newcommand{\pose}[0]{\boldsymbol{\theta}}
\newcommand{\shape}[0]{\boldsymbol{\beta}}
\newcommand{\mesh}[0]{\mathcal{M}}
\newcommand{\loss}[0]{\mathcal{L}}
\newcommand{\expectation}[0]{\mathbb{E}}
\newcommand{\encoder}[0]{\mathcal{E}}

\newcommand{\norm}[1]{\left\lVert#1\right\rVert}
\newcommand{\cbest}[1]{\cellcolor{limegreen}\textbf{#1}}
\newcommand{\csecond}[1]{\cellcolor{myyellow}#1}
\definecolor{limegreen}{HTML}{badc58}
\definecolor{myyellow}{HTML}{f6e58d}
\newcommand{\best}[1]{\colorbox{limegreen}{\textbf{#1}}}
\newcommand{\second}[1]{\colorbox{myyellow}{#1}}

\definecolor{markgreen}{HTML}{27ae60}
\definecolor{markred}{HTML}{e74c3c}
\newcommand{\cmark}{\textcolor{markgreen}{\ding{51}}}%
\newcommand{\xmark}{\textcolor{markred}{\ding{55}}}%
\newcolumntype{R}[2]{%
    >{\adjustbox{angle=#1,lap=\width-(#2)}\bgroup}%
    l%
    <{\egroup}%
}
\newcommand*\rot{\multicolumn{1}{R{20}{1em}}}

\renewcommand{\paragraph}[1]{\vspace{0.2mm}\noindent\textbf{#1}\:}

\usepackage[pagebackref,breaklinks,colorlinks,citecolor=eccvblue]{hyperref}

\usepackage{orcidlink}

\begin{document}

\title{Instant 3D Human Avatar Generation using Image Diffusion Models\vspace{-3mm}} 

\titlerunning{Instant 3D Human Avatar Generation}

\author{Nikos Kolotouros\orcidlink{https://orcid.org/0000-0003-4885-4876} \and
Thiemo Alldieck\orcidlink{https://orcid.org/0000-0002-9107-4173} \and
Enric Corona\orcidlink{https://orcid.org/0000-0002-4835-1868} \and
Eduard Gabriel Bazavan \and
Cristian Sminchisescu\orcidlink{https://orcid.org/0000-0001-5256-886X}}

\authorrunning{N.~Kolotouros et al.}

\institute{Google Research\footnote[1]{Now at Google DeepMind.}\\
\email{\{kolotouros,alldieck,egbazavan,enriccorona,sminchisescu\}@google.com}}
\maketitle

\begin{figure}
    \vspace{-10mm}
    \centering
    \includegraphics[width=\textwidth]{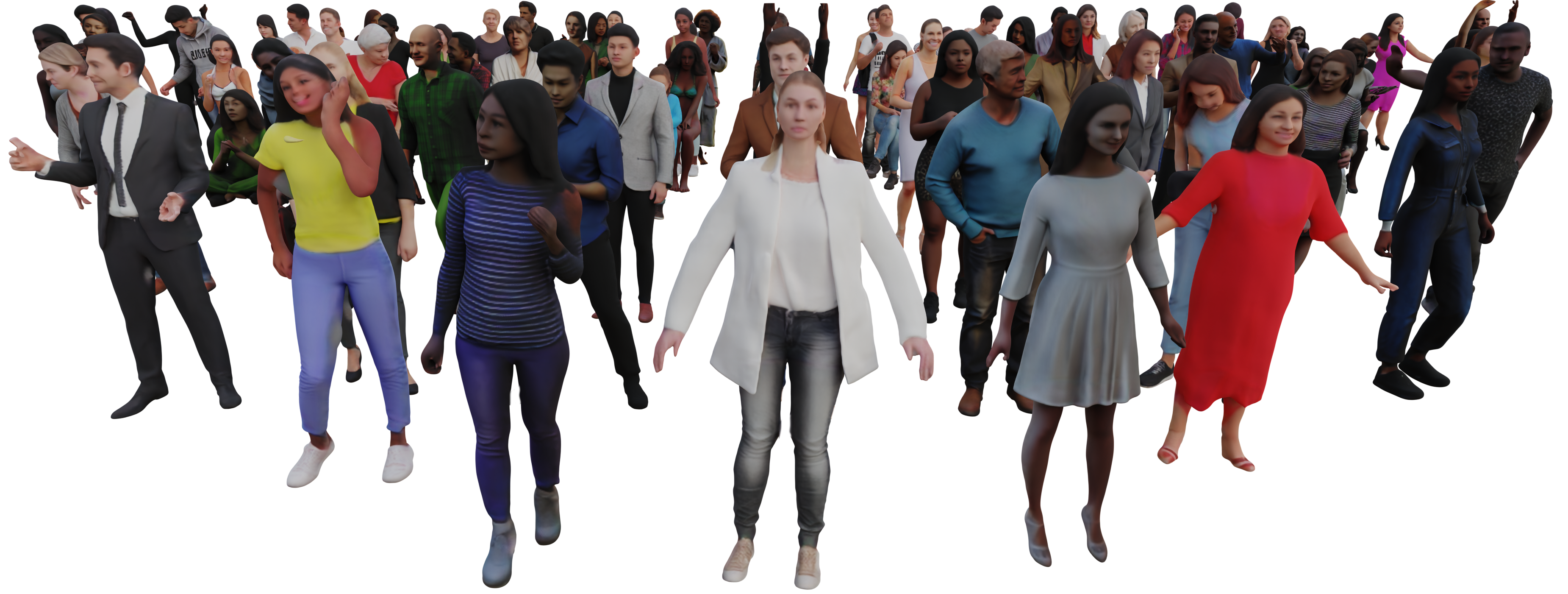}
    \caption{%
    We present \textbf{\method}, a new method for the automatic generation of 3D human assets. \method can generate rigged 3D models from text or from single images and has control over body pose and shape. In this example, we show 77 models generated from various text prompts in 12 minutes on a single GPU.
    }
    \label{fig:teaser}
\end{figure}

\vspace{-8mm}
\input{sections/abstract}
\input{sections/introduction}

\input{sections/related}
\begin{figure}[t]
    \centering
    \includegraphics[width=\textwidth]{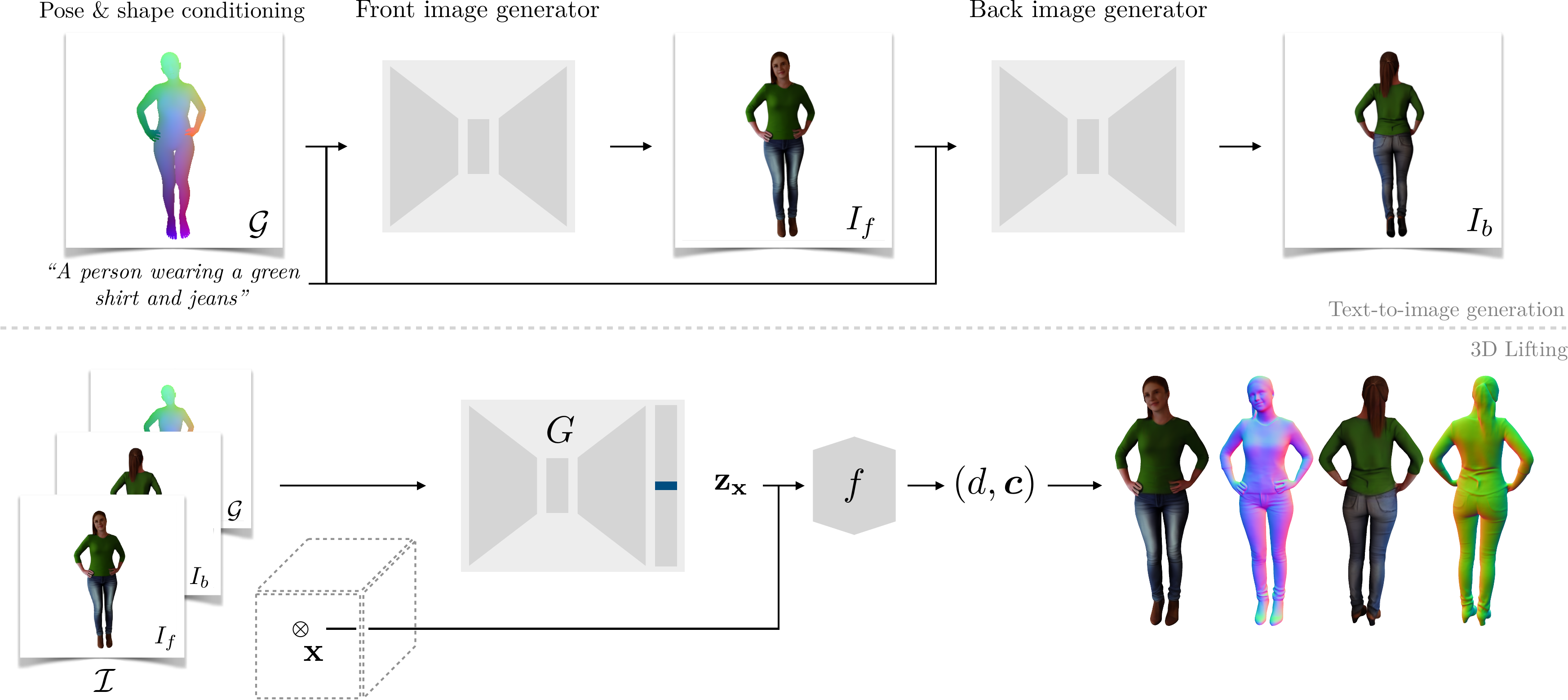}
    \caption{\textbf{\method method.} %
    (Top) \method builds on the capacity of text-to-image models to generate highly detailed and diverse input images. First, a Latent Diffusion network takes a text prompt and a target body pose and shape $\mathcal{G}$, and generates a highly detailed front image $I_f$ of a person. Next, a second network generates a consistent back view $I_b$ in the same pose and clothing. 
    (Bottom) 
    We perform pixel-aligned 3D reconstruction given the generated front/back views $I_f, I_b$  and optionally the given 3D body pose and shape $\mathcal{G}$. This decoupling enables the generation of 3D avatars from text, images or a combination of the two. %
    }
    \label{fig:method}
\end{figure}
\input{sections/method}

\input{sections/experiments}

\input{sections/conclusion}
\appendix
\section*{\Large Supplementary Material}

The Supplementary Material contains additional implementation details and experiments that were not included in the main paper due to space constraints. 
Additional results can be found on our project website: \\
\hyperlink{https://www.nikoskolot.com/avatarpopup/}{https://www.nikoskolot.com/avatarpopup/}.

\input{sections/supp_content}

\newpage
\onecolumn

\bibliographystyle{splncs04}
\bibliography{main}
\end{document}

%% file: sections/abstract.tex
\begin{abstract}
\vspace{-5mm}
We present \method, a method for fast, high quality 3D human avatar generation from different input modalities, such as images and text prompts and with control over the generated pose and shape.
The common theme is the use of diffusion-based image generation networks that are specialized for each particular task, followed by a 3D lifting network.
We purposefully decouple the generation from the 3D modeling which allow us to leverage powerful image synthesis priors, trained on billions of text-image pairs.
We fine-tune latent diffusion networks with additional image conditioning for image generation and back-view prediction, and to support qualitatively different multiple 3D hypotheses.
Our partial fine-tuning approach allows to adapt the networks for each task without inducing catastrophic forgetting.
In our experiments, we demonstrate that our method produces accurate, high-quality 3D avatars with diverse appearance that respect the multimodal text, image, and body control signals.
Our approach can produce a 3D model in as few as 2 seconds, a \emph{four orders of magnitude speedup} \wrt the vast majority of existing methods, most of which solve only a subset of our tasks, and with fewer controls.
\method enables applications that require the controlled 3D generation of human avatars at scale. The project website can be found at \hyperlink{https://www.nikoskolot.com/avatarpopup/}{https://www.nikoskolot.com/avatarpopup/}.
\end{abstract}

%% file: sections/introduction.tex
\section{Introduction}
We present \method, a method for instant generation of rigged full-body 3D human avatars, with multimodal controls in the form of text, images, and/or human pose and shape. 
The remarkable recent progress in image synthesis \cite{sohl2015deep, ho2020denoising, dhariwal2021diffusion, ramesh2021zero, saharia2022photorealistic, rombach2022high} acted as a catalyst for a wide range of media generation applications.
In just a few years, we have witnessed rapid developments in video generation \cite{ho2022imagen, villegas2022phenaki, zeng2023make, kondratyuk2023videopoet, bar2024lumiere}, audio synthesis \cite{oord2016wavenet, zhang2023audio} or text-to-3D object generation \cite{poole2022dreamfusion, raj2023dreambooth3d, liu2023zero, liu2024one, sun2023dreamcraft3d}, among others. 
Pivotal to success of all these methods is their probabilistic nature,  however this requires very large training sets.
While inspiring efforts have been made \cite{deitke2023objaverse}, training set size is still a problem in many domains, and particularly for 3D.
In an attempt to alleviate the need for massive 3D datasets, DreamFusion \cite{poole2022dreamfusion} leverages the rich priors of text-to-image diffusion models in an optimization framework. 
The influential DreamFusion ideas were also quickly adopted for 3D avatar creation \cite{kolotouros2023dreamhuman,liao2023tada}, a field previously dominated by image- or video-based reconstruction solutions \cite{saito2019pifu,alldieck2018video,saito2020pifuhd,alldieck2022phorhum}.
Text-to-avatar methods enabled novel creative processes, but came with a significant drawback. While image-based methods typically use pretrained feed-forward networks and create outputs in seconds, existing text-to-avatar solutions are optimization-based and take minutes to several hours to complete, per instance.

In this paper, we are closing this gap and present, for the first time, a methodology for \emph{instant}, text-controlled, rigged, full-body 3D human avatar creation.
Our \method is purely feed-forward, can be conditioned on images and textual descriptions, allows fine-grained control over the generated body pose and shape, can generate multiple hypotheses, and runs in 2-10 seconds per instance.

Key to success is our pragmatic decoupling of the two stages of probabilistic text-to-image generation and 3D lifting.
Decoupling 2D generation and 3D lifting has two major advantages: (1) We can leverage the power of pretrained text-to-image generative networks, which have shown impressive results in modeling complex conditional distributions. Trained with large training sets of images, both generation quality and diversity are very high. (2) We alleviate the need for very large 3D datasets required to train state-of-the-art  generative 3D models.
Our method generates diverse plausible image configurations that contain rich enough information that can be lifted in 3D with minimal ambiguity.
In other words, we distribute the workload between two expert systems: a pretrained probabilistic generation network fine-tuned for our task to produce front and back image views of the person, and a state-of-the-art unimodal, feed-forward image-to-3D model that can be trained using comparably small datasets.

Our proposed decoupling strategy allows us to maximally exploit available data sources with different levels of supervision. We first fine-tune a pretrained Latent Diffusion network to generate images of people based on textual descriptions and with additional control over the desired pose and shape.
This step does not require any ground truth 3D data for supervision and enables scaling our image generator to web scale data of images of people in various poses.
Next, we leverage a small-scale dataset of scanned 3D human assets and fine-tune a second latent diffusion network to learn the distribution of back side views conditioned on a front view image of the person. We optionally also condition on a textual description that can naturally complement the evidence available in the front view image.
Furthermore, we propose a novel fine-tuning strategy that prevents overfitting to the new datasets.
Finally, we design and train a 3D reconstruction network that predicts a textured 3D shape in the form of an implicit signed distance field given the pair of front and back views and optionally 3D body signals.
The resulting cascaded method, which we call \method supports a wide range of 3D generation and reconstruction tasks:
First, it enables fast and interactive 3D generation of assets at scale, see \cref{fig:teaser}.
Second, we can repurpose parts of the cascade for image-based 3D reconstruction at state-of-the-art quality.
Finally, we demonstrate how \method can be used for creative editing tasks exemplified in 3D virtual try-on with body shape preservation.
To summarize, our main contributions are:
\begin{itemize}
    \item A method for controllable 3D human avatar generation, based on multimodal text, pose, shape and image input signals, that outputs a detailed human mesh instance in 2-10 seconds.
    \item We propose a simple yet effective way to fine-tune pretrained diffusion models on small-scale datasets, without inducing catastrophic forgetting.
    \item While not our primary goal, our approach achieves state-of-the art results in single-image 3D reconstruction and enables 3D creative editing applications.
\end{itemize}

%% file: sections/related.tex
\section{Related Work}

\begin{table}[t]
\centering
\setlength{\tabcolsep}{12pt}
\resizebox{\linewidth}{!}{%
    \begin{tabular}{cccccccc}
    \rot{Control from Text} & \rot{Control from Image} & \rot{Body Pose/Shape Control} & \rot{Generates Geometry} & \rot{Generates Texture} &
    \rot{Editable} & \rot{Runtime} & \\ 
    \hline
    \cmark & \cmark & \cmark & \cmark & \cmark & \cmark/\xmark & several hours & Optimization-based text-to-3D~\cite{kolotouros2023dreamhuman, liao2023tada, hong2022avatarclip} \\
    \xmark & \cmark & \xmark & \cmark & \cmark/\xmark & \xmark & seconds & Image-to-3D \cite{saito2019pifu, saito2020pifuhd, alldieck2022phorhum, alldieck2019learning} \\
    \cmark & \xmark & \cmark & \cmark & \xmark & \xmark & minutes & Hybrid \cite{kim2023chupa} \\
    \xmark & \xmark & \cmark & \cmark & \cmark & \xmark & seconds & 3D GANs~\cite{hong2023evad, dong2023ag3d} \\
    \hline
    \cmark &\cmark & \cmark & \cmark & \cmark & \cmark & seconds & \textbf{AvatarPopUp (Ours)} \\
\end{tabular}}
\vspace{1mm}
\caption{AvatarPopUp generates 3D assets with texture from text prompts or input images of a target subject and can be controlled with body pose and shape. 
In contrast to baselines that require up to hours per prompt, our model takes under five seconds and can de facto by used in interactive applications.
In the experimental section we demonstrate the large diversity of the model, and applications in cloth editing.
}
\label{tab:rw}
\vspace{-7mm}
\end{table}

Table \ref{tab:rw} summarizes the characteristics of AvatarPopUp in comparison to previous work, along several important property axes.

\paragraph{Text-to-3D generation.} 
The success of text-to-image models~\cite{dalle, stablediffusion, imagen} was quickly followed by a significant amount of work on text-to-3D content generation~\cite{poole2022dreamfusion, lin2023magic3d, lorraine2023att3d,tang2023dreamgaussian}.
Due to limited training data, methods typically use optimization approaches, where a neural representation~\cite{nerf} is optimized per instance by minimizing a distillation loss~\cite{poole2022dreamfusion} derived from large text-to-image models.
This idea has been extended to generate human avatars~\cite{huang2024tech, liao2023tada, kolotouros2023dreamhuman, gong2024text2avatar, jiang2023avatarcraft, xu2023seeavatar, zhang2023text, zhang2023avatarverse, zhang2023avatarstudio,  wang2023disentangled} or heads~\cite{han2024headsculpt, liu2023headartist, lei2023diffusiongan3d} 
enabling the text-based creation of 3D human assets that are diverse in terms of shape, appearance, clothing and various accessories.
In these works, the optimization process is often regularized using a 3D body model~\cite{kolotouros2023dreamhuman, liao2023tada, zhang2023text}, which also enables animation.
However, such approaches generally take hours per instance, and rendering is slow.
With the appearance of Gaussian Splatting~\cite{gaussiansplatting}, other works~\cite{liu2023humangaussian,zhao2024psavatar,abdal2023gaussian} reduced rendering time at the expense of accurate geometry.
In any case, creating an avatar still takes a significant amount of time, making such methods unsuitable for interactive applications. 
In this work we propose an alternative direction, which also builds upon the success of text-to-image models and combines them with 3D reconstruction pipelines.
Also related are 3D human generation methods. AG3D \cite{dong2023ag3d} and EVA3D \cite{hong2023evad} are GAN-based methods learned from 2D data, that allow to sample 3D humans anchored in a 3D body model.
CHUPA \cite{kim2023chupa} generates dual normal maps based on text and then fits a body model to obtain a full 3D  representation.
While generation is similar in spirit to our method, CHUPA requires optimization per instance and does not generate texture.

\paragraph{Photorealistic 3D Human Reconstruction.}
Our framework generates 3D human assets and builds on top of state-of-the-art 3D reconstruction techniques. This has been widely explored in the past and can be roughly categorized by its use of explicit or implicit representations. 
An important line of work leverages 3D body models~\cite{smpl, smplx, xu2020ghum} and reconstructs their associated parameters,  in some cases extended with vertex offsets to represent some clothing and hair detail~\cite{alldieck2019learning, alldieck2018detailed, alldieck2018video, tex2shape, onizuka2020tetratsdf, zhu2019detailed}.
Other efforts have considered voxels~\cite{varol2018bodynet, zheng2019deephuman}, depth maps~\cite{gabeur2019moulding} and more recently implicit representations~\cite{saito2019pifu, saito2020pifuhd, alldieck2022phorhum, xiu2022icon, xiu2023econ, zheng2021pamir, yang2021s3, integratedpifu, he2021arch++, huang2020arch, corona2023s3f, diffhuman}.
Being topology free, the latter allow the representation of loose clothing more easily.
They typically provide more detail and enable high-resolution reconstruction, often conditioned on local pixel-aligned features~\cite{saito2019pifu}.
On the other hand, these methods yield reconstructions with no semantic labels that cannot be easily animated.
To solve this problem, some work combined body models with implicit representations~\cite{huang2020arch, he2021arch++, corona2023s3f, xiu2022icon, xiu2023econ, huang2024tech}, but this is prone to errors when the pose is noisy at inference time.
In contrast, we drive the synthesis process with guidance from an input body model -- sampled or estimated -- so that the generated image is well aligned with the body prior. As we show in \cref{sec:experiments}, this allows rigging the 3D avatar without post-processing and natively supports 3D animation.

Given a single input image of a person, previous work aims to generate realistic reconstructions for the non-visible parts.
However, this often leads to blurry textures and lack of geometric detail, \eg no wrinkles.
Some methods~\cite{saito2019pifu, saito2020pifuhd, xiu2022icon, xiu2023econ} generate back normal maps to enhance details, or consider probabilistic reconstructions \cite{diffhuman, albahar2023humansgd}. However, all these methods cannot be prompted from text or other modalities and still yield limited diversity. 
In contrast, we guide the synthesis process by means of generated front and back images, yielding high-quality 3D reconstructions.
Another challenge in previous work is limited training data. Most prior methods rely on a few hundred 3D scans, due to the pricey and laborious process of good quality human capture. We alleviate the need for large scale 3D training data by proposing a framework that can quickly generate humans with a given clothing, pose and shape.

%% file: sections/method.tex
\section{Method}

We learn a distribution $p(X | c)$ of textured 3D shapes $X$ conditioned on a collection of signals $c$ that we factorize as follows
\begin{equation}
    p(X | c) = \int \int p(X | I_f, I_b, c) \cdot p(I_b | I_f, c) \cdot p(I_f | c) dI_f dI_b,
    \label{eq:probability}
\end{equation}
where $p(X | I_f, I_b, c)$ is the probability of 3D shape $X$ given $c$ and front and back image observations $I_f$ and $I_b$ respectively, $p(I_b | I_f, c)$ is the probability of the back view image given the front image $I_f$ and conditioning signals $c$, and $p(I_f | c)$ the conditional probability of front view images of the person given $c$.

Computing the integral in \eqref{eq:probability} is intractable, but our goal is to generate samples from the distribution rather than expectations.
To do so, we employ ancestral sampling.
We first sample a front view $I_f$ given $c$, we then sample a back view $I_b$ given $I_f$ and $c$, and last we sample the 3D reconstruction based on the entire context.
In practice, $p(I_f | c)$ and $p(I_b | I_f, c)$ are implemented using Latent Diffusion models, whereas $p(X | I_f, I_b, c)$ is a unimodal, neural implicit field generator.

In the case of single-image 3D reconstruction the conditioning signal $c$ is $I_f$, and consequently we can omit the first step. For text-based generation, $c$ is a text prompt describing the appearance of the person together with a signal encoding the body pose and shape. %
The conditioning information $c$ may be extended with additional signals, as in the case of 3D editing, \cf \cref{sec:experiments}.

\subsection{Controllable Text-to-Image Generator}
Recent advances in diffusion-based text-to-image generation networks \cite{rombach2022high} have enabled synthesizing high-quality images given only a text prompt as input.
However, for certain use cases, such as human generation, it is difficult to inject fine-grained, inherently continuous forms of control, like the 3D pose of people or their precise body shape proportions, when generating with text alone.

Inspired by ControlNet \cite{zhang2023controlnet}, we propose to add simultaneous control over body pose and shape by augmenting a pretrained Latent Diffusion network with an additional image input that jointly encodes both modalities.
For control we use GHUM \cite{xu2020ghum} but other models (\eg \cite{smpl}) can be used. Specifically given 3D pose and shape parameters $\pose$ and $\shape$ respectively, we render the corresponding mesh $\mesh = \mathrm{GHUM}(\pose, \shape)$ using GHUM's template coordinates and posed vertex locations as 6D vertex colors, obtaining a dense, pixel-aligned pose- and shape-informed control signal $\mathcal{G}$.

To fine-tune the network, we generate a dataset of images of people with corresponding GHUM 3D pose and shape parameters and text annotations. This dataset is comprised of a set of scanned assets \cite{renderpeople} that are rendered from different viewpoints, as well as a set of real images scraped from the web. For the synthetic part of the dataset, the pose and shape parameters are obtained by fitting GHUM to 3D scans. We additionally use real images to which we fit GHUM using keypoint optimization in the style of \cite{kolotouros2019learning}. For all images we obtained text annotations using an off-the shelf image captioning system \cite{xi2022pali} by prompting it to describe the clothing of the people in the image. Since we are interested in generating 3D human assets, we additionally mask out the background in all images, and train the network to output segmented images. This makes the downstream 3D reconstruction task easier, and improves the reconstruction quality because it encourages the network to focus on  human appearance, rather than allocating capacity to model complex backgrounds.

We want to exploit the rich priors learned by text-to-image foundation models by fine-tuning a Latent Diffusion model \cite{rombach2022high} with the dense GHUM rendering as an additional input. For fine-tuning we propose a simpler and more lightweight method than a standard ControlNet. We pad the weights of the input convolutional layer with additional channels initialized with zeros, and we then fine-tune only the weights of the convolutional layers of the encoder network.
All the decoder and attention layers are kept frozen.
With this simple strategy, even though our model is trained on a relatively small set of images, it can generalize to unseen types of clothing.
At the same time, our strategy is more practical than training a ControlNet, which involves keeping in separate copy of the original network weights in memory, and thus enables fine-tuning large models with moderate hardware utilization.
We optimize the encoder of the diffusion model by minimizing the simple variant of the diffusion loss \cite{ho2020denoising}
\begin{equation}
    \loss (\boldsymbol{\psi}_{\text{enc}}) =  \expectation_{\encoder(x), \epsilon, t, \tau, \mathcal{G}}
    \norm{\epsilon -  \epsilon_{\boldsymbol{\psi}} (z_t, t, \tau, \encoder(\mathcal{G}) )},
\end{equation}
where $t \in \{1, \dots, T\}$ is the diffusion time step, $\epsilon \sim \mathcal{N}(0, I)$ the injected noise $z_t = \alpha_t \encoder(x) + \nu_t$ is the noisy image latent, $\tau$ is the text encoding, $\encoder(\mathcal{G})$ the latent encoding of the dense GHUM signal, and $\boldsymbol{\psi}_{\text{enc}}$ the encoder subset of the denoising UNet parameters $\boldsymbol{\psi}$.

\subsection{Back View Generation}
\begin{figure}[t]
    \centering
    \begin{subfigure}{\linewidth}
    \centering
    \includegraphics[width=.22\textwidth]{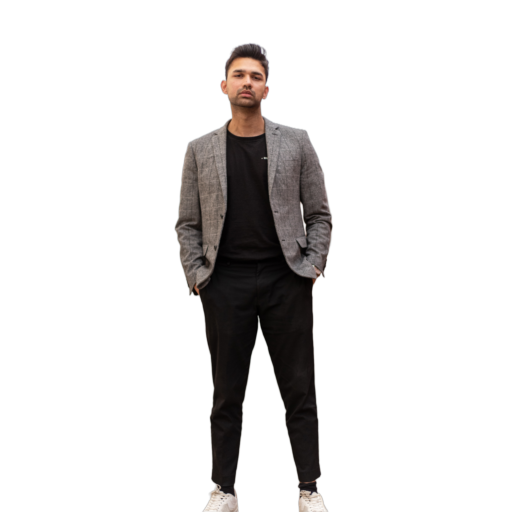}
    \includegraphics[width=.22\textwidth]{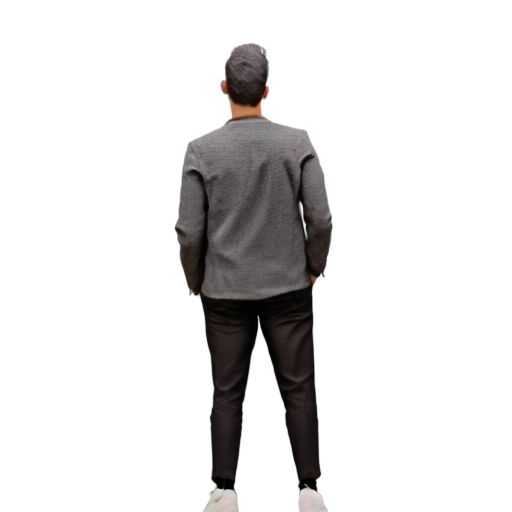}
    \includegraphics[width=.22\textwidth]{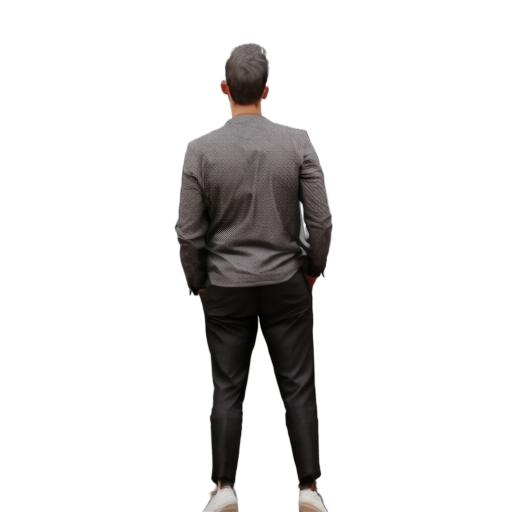}
    \includegraphics[width=.22\textwidth]{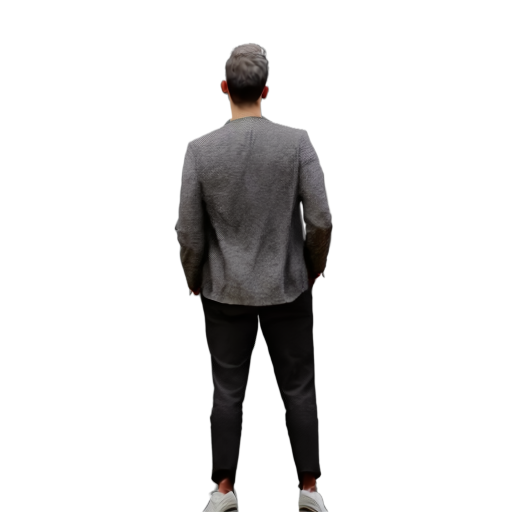}
    \vspace{1.0mm}
    \caption*{\scriptsize{{No text conditioning}}}
    \vspace{1.0mm}
    \end{subfigure}\\

    \centering
    \begin{subfigure}{.22\textwidth}
    \includegraphics[width=\textwidth]{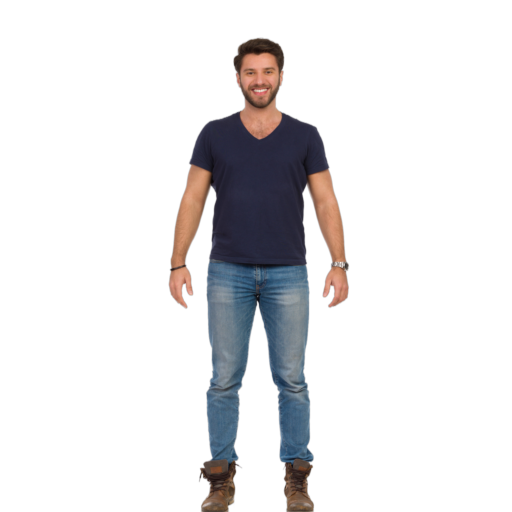}
    \caption*{}
    \label{fig:short-a}
    \end{subfigure}
    \begin{subfigure}{.22\textwidth}
    \includegraphics[width=\textwidth]{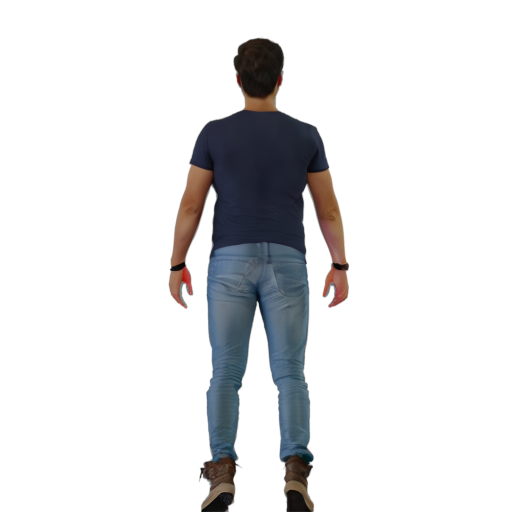}
    \caption*{}
    \label{fig:short-a}
    \end{subfigure}
    \begin{subfigure}{.22\textwidth}
    \includegraphics[width=\textwidth]{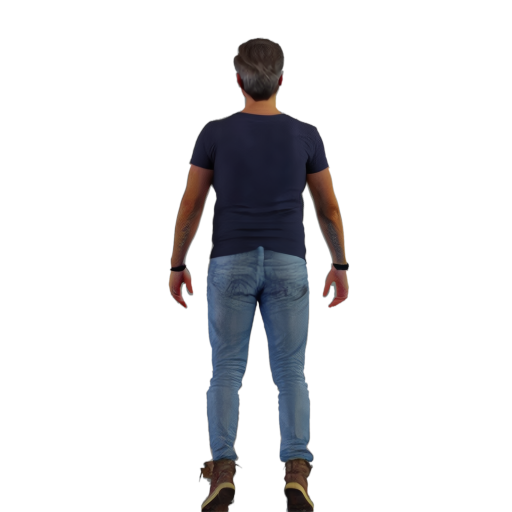}
    \caption*{+ ``{gray hair}''}
    \label{fig:short-a}
    \end{subfigure}
    \begin{subfigure}{.22\textwidth}
    \includegraphics[width=\textwidth]{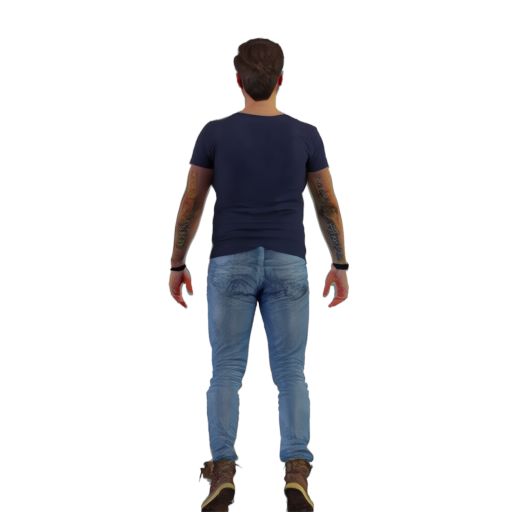}
    \caption*{+ ``{with a tattoo}''}
    \label{fig:short-a}
    \end{subfigure}
    \vspace{3.0mm}
    \caption{\textbf{Diverse back view hypotheses.} Conditioned on the front view, our method is able to generate diverse plausible back views of the person, with different hairstyles, wrinkle patterns, or lighting. Our network can also be controlled with text (second row), to add fine-grained detail to our generated back-side views.}
    \label{fig:back_side_examples}
\end{figure}

\vspace{-1mm}
One could try to lift the generated front views from the previous stage to 3D directly by applying a single-image 3D reconstruction method like PHORHUM \cite{alldieck2022phorhum}. However, because of the inherent ambiguity of the problem this will result in significant loss of geometric detail and blurry textures for the non-visible body surfaces. To avoid this, we propose to fine-tune again a latent diffusion network with the same strategy as in the previous section. This time the additional image conditioning is a front view and optionally a text prompt, and we train the network to learn the distribution of back views conditioned on the front view. The additional text prompt can be used in cases where it is desired to additionally guide the generation by very specific properties.
\cref{fig:back_side_examples} shows different back sides sampled from the conditional distribution. We also show that the additional text inputs are useful in modulating certain parts of the generation that are not immediately deducible from the front image, such as hairstyles or specific patterns.

\subsection{3D Reconstruction Model}
\vspace{-1mm}
Our 3D reconstruction network is inspired by PHORHUM \cite{alldieck2022phorhum}, and our design choices are informed by the limitations of typical single-image 3D reconstruction methods.
Specifically, given a collection of input image signals $\mathcal{I} = \{I_f, I_b, \mathcal{G}\}$, we first concatenate them, and then use a convolutional encoder $G$ to compute a pixel-aligned feature map $G(\mathcal{I})$.
The 3D body control signal $\mathcal{G}$ is optional and may be omitted, \eg for single-image reconstruction.
Then, each point $\mathbf{x} \in \mathbb{R}^3$ in the scene gets projected on this feature map to get pixel-aligned features $\mathbf{z}_\mathbf{x} = g(\mathcal{I}, \mathbf{x}; \pi) = b(G(\mathcal{I}), \pi(\mathbf{x}))$ using interpolation, where $b(\cdot)$ is the bilinear sampling operator and $\pi(\mathbf{x})$ is the pixel location of the projection of $\mathbf{x}$ using the camera $\pi$.
These pixel aligned features are then concatenated with a positional encoding $\gamma(\mathbf{x})$ of the 3D point and are fed to an MLP $f$ that outputs the signed distance from the surface $d$ as well as surface color $\boldsymbol{c}$. Finally, the 3D shape $\mathcal{S}$ is represented as the zero-level-set of $d$
\begin{equation}
    \mathcal{S}(\mathcal{I}) = \{\mathbf{x} \in \mathbb{R}^3 | f(g(I, \mathbf{x}; \pi), \gamma(\mathbf{x})) = (0, \boldsymbol{c})\}.
\end{equation}
$\mathcal{S}$ can be transformed to a mesh directly, using Marching Cubes \cite{marching_cubes}.

\subsection{Animation of Generated Avatars}
\vspace{-1mm}
Our method can generate diverse 3D avatars with various poses, shapes and appearances. Optionally, we may leverage the conditioning body model to rig the estimated 3D shape.
As a result of our conditioning strategy, 3D avatars and the conditional body model instances are aligned in 3D. 
This allows us to anchor the reconstructed 3D shape on the body model surface \cite{bazavan2021hspace} and re-pose or re-shape it accordingly.
Alternatively, we may also transfer just the LBS skeleton and weights from the body model to the scan. This enables importing and animating the generated 3D assets in various rendering engines.
See \cref{fig:animation} and Sup.\ Mat.\ for examples and videos of animations of the generated 3D assets.

%% file: sections/experiments.tex
\section{Experiments}
\label{sec:experiments}

\paragraph{\bf Data.} 
We use meshes from RenderPeople~\cite{renderpeople} for training as well as our own captured data, totaling in $\sim$10K scans with diverse poses, body shapes, and clothing styles. 
We render each scan with randomly sampled HDRI background, random cloth color augmentations, and lighting using Blender~\cite{blender}. During this process, we render both front and back views used to train the different stages of our model. For the front image generation network we also use a set of 10K real images on which we fitted the GHUM model using 2D keypoints.  
For testing we defined a split based on subject identity and held out $\sim$1K scans.

We provide results for 2 different versions of our model. The standard quality model is generated in $2$ seconds by running 5 DDIM \cite{song2020denoising} steps during inference and Marching Cubes at $256^3$ resolution. The high quality model is generated in $10$ seconds using 50 DDIM steps and Marching Cubes at $512^3$ resolution. The timings were recorded on a single $40$ GB A100 GPU. Unless otherwise stated, all results we report are obtained using the high quality model.

\paragraph{\bf Metrics and Baselines.} 
We compare our method numerically for two different problems. First we consider the task of text-to-3D human generation, where we sample 100 different text prompts and compare against representative text-to-3D generation methods. For numerical comparisons we evaluate the text-image alignment using CLIP \cite{radford2021learning}. Specifically, we use the retrieval accuracy using CLIP, as proposed in \cite{poole2022dreamfusion}. We additionally show qualitative results.

Second, we validate the performance of our 3D reconstruction component against state-of-the-art methods~\cite{saito2019pifu, saito2020pifuhd, xiu2022icon, xiu2023econ, alldieck2022phorhum, huang2024tech} considering both geometry and texture. Pixel-aligned image features dominate in recent work~\cite{saito2019pifu, saito2020pifuhd, alldieck2022phorhum}, but some methods proposed to combine them with body models~\cite{xiu2022icon, xiu2023econ}, which offer advantages in being able to animate reconstructions. We also rely on pixel-aligned features, yet propose a method that inherently enables animation. 
We run \method by generating the back side of the subject and applying the reconstruction network. To evaluate 3D geometry, we report bi-directional Chamfer distance $\times 10^{-3}$, Normal Consistency (NC $\uparrow$), and Volumetric Intersection over Union (IoU $\uparrow$) after ICP alignment. However, these metrics do not necessarily correlate with good visual quality, \eg Chamfer distance is minimized by smooth, non-detailed geometry. To measure the quality of reconstructions, we additionally report FID scores~\cite{fid} of the front/back views for both geometry and texture. 

\subsection{Text-to-3D Generation}
\begin{figure}[t]
    \centering
    \includegraphics[width=.9\textwidth]{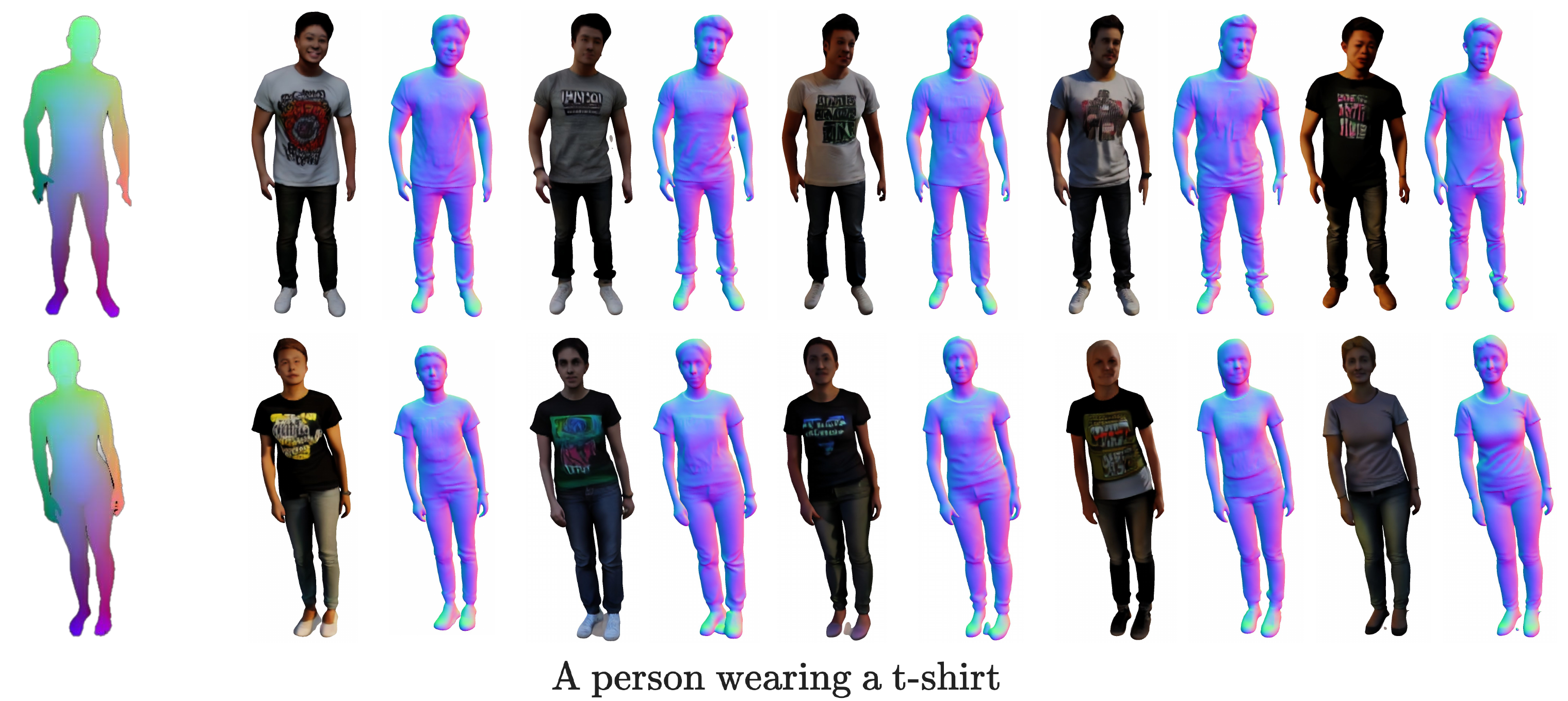} 
    \caption{\textbf{Diversity of our 3D generation.} For the same text prompt and the same pose and shape conditioning, our model can generate a diverse set of 3D avatars that respect both the text and the 3D body controls.}
\label{fig:diversity}
\end{figure}

In \cref{fig:diversity} we generate different avatars given the same text prompt and driving poses. We can see that our model is able to create a very diverse set of assets, a property not observed in the previous text-to-3D generation methods \cite{kolotouros2023dreamhuman}.

\begin{figure}[t]
    \centering
    \includegraphics[width=.9\textwidth]{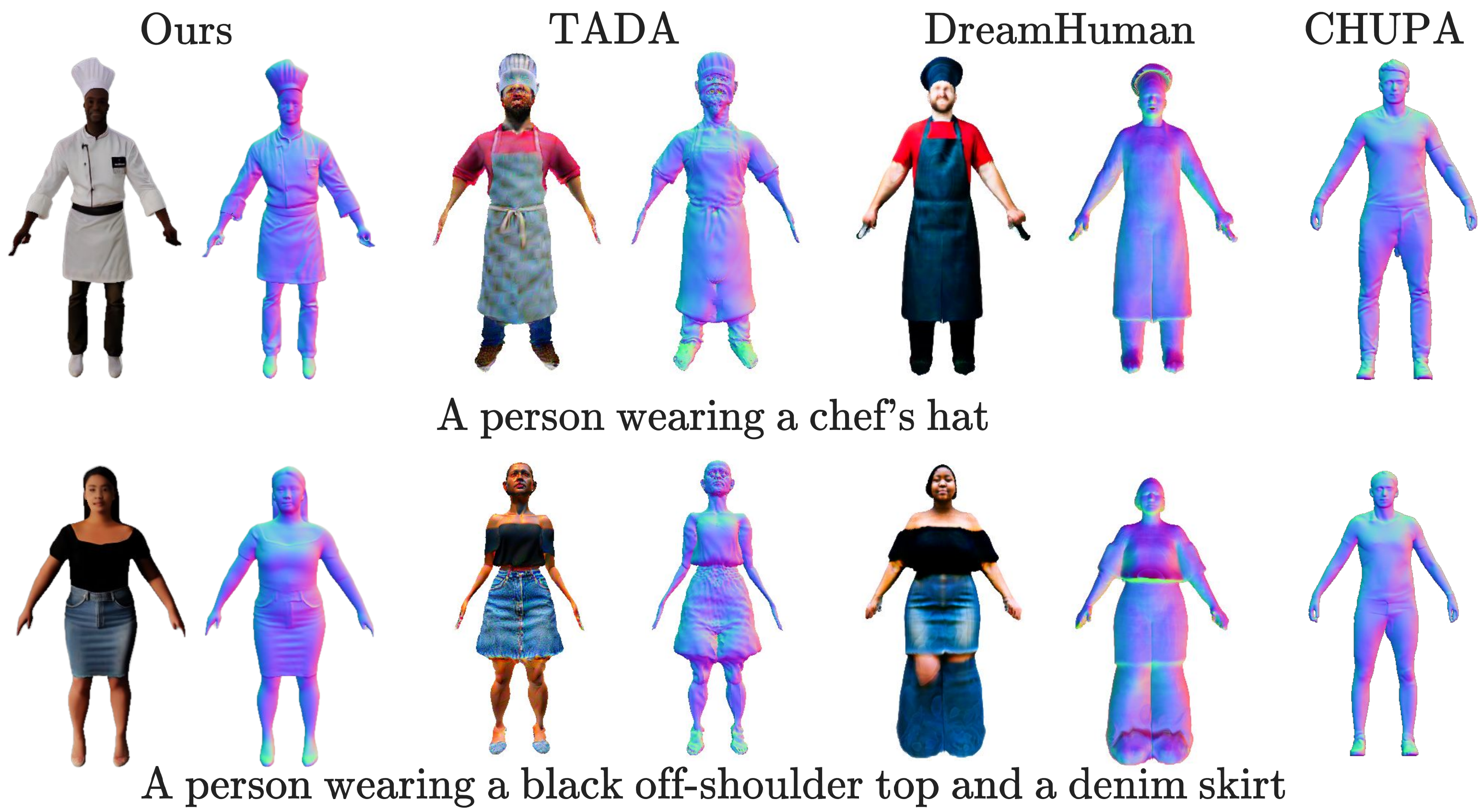}
    \caption{\textbf{Comparisons with text-to-3d human generation methods.} Our method generates high quality results that respect the text prompt well, at a fraction of the others' runtime, \cf \cref{tab:numerical_eval_image}. TADA's results appear unnatural; DreamHuman failed for one subject and produces oversaturated colors; CHUPA failed to respect the prompt.}
\label{fig:gen3d_comp}
\end{figure}

In \cref{tab:numerical_eval_gen} we use CLIP to evaluate our model against other text-to-3D generation methods. In general, CLIP-based metrics are not indicative of the generated image quality, because they only consider the alignment with the text, and often over-saturated images with extreme details tend to have high CLIP scores. To further demonstrate that our method generates higher quality avatars, we also include a qualitative comparison in \cref{fig:gen3d_comp}.

\begin{table}[t]
\vspace{-0.3cm}
    \centering
    \resizebox{0.7\linewidth}{!}{%
        \def\arraystretch{1.1}%
        \setlength{\tabcolsep}{3pt}
        \begin{tabular}{cc|cc|c|l}
            \multicolumn{2}{c|}{Color} &  \multicolumn{2}{c|}{Color}  & &  \\
            R-Prec. $\uparrow$ & Top-3 $\uparrow$ & R-Prec. $\uparrow$  & Top-3 $\uparrow$ & Runtime $\downarrow$ & \\
            \hline
            \cbest{0.68} & \cbest{0.92} & 0.04 & \cbest{0.25} & 8h& DreamHuman \cite{kolotouros2023dreamhuman}\\
            0.56 & \csecond{0.82} & 0.03 & 0.15 & 3h & TADA \cite{liao2023tada}\\
            -- & -- & 0.03 & 0.08 & 3m & CHUPA \cite{kim2023chupa}\\ \hline
            0.58 & 0.73 & \csecond{0.08} & \csecond{0.17} & \cbest{ $\sim$2s} & Ours\\
            \csecond{0.62} & 0.77 & \cbest{0.11} & \csecond{0.17} & \csecond{ $\sim$10s} & Ours (high quality)\\
        \end{tabular}
    }
    \vspace{2mm}
    \caption{\textbf{Numerical comparisons with other text-to-3d human generation methods.}  We mark the \best{best} and \second{second best} results. Our method allows to trade-off speed and quality and we report results for two different settings.}
    \label{tab:numerical_eval_gen}
    \vspace{-7mm}
\end{table}

\subsection{Single-image 3D Reconstruction}
\begin{figure}
\vspace{-0.3cm}
    \centering
    \includegraphics[width=\textwidth]{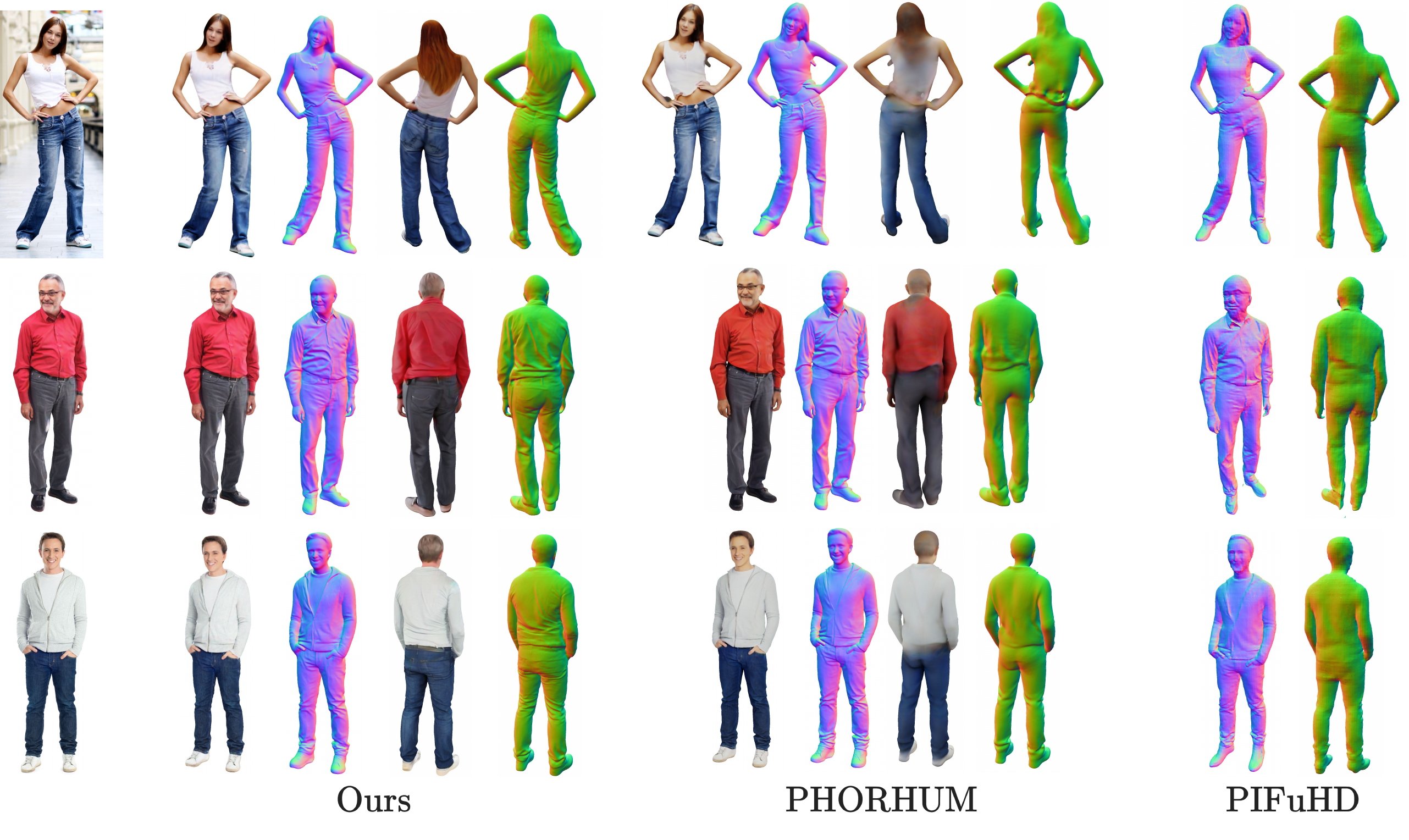} 
    \caption{\textbf{Qualitative comparisons with state-of-the-art single image 3D reconstruction methods.} Our method produces front color and normals on par with state-of-the-art and much more detailed back view hypotheses.}
\label{fig:imageto3d}
\end{figure}

\begin{figure}
\vspace{-0.3cm}
    \centering
    \includegraphics[width=\textwidth]{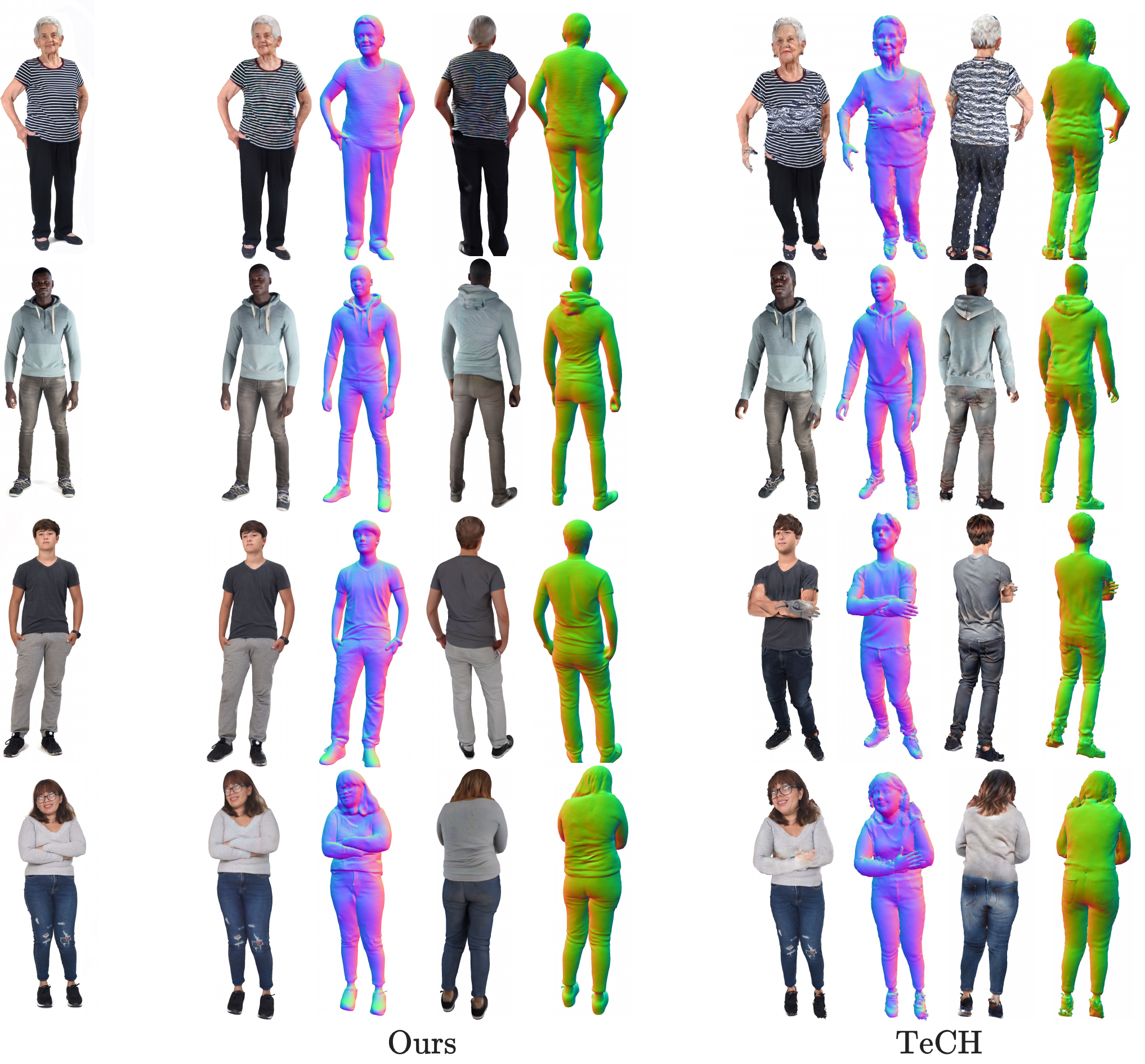} 
    \caption{\textbf{Additional comparisons with TeCH.} TeCH is optimization-based and takes several hours to complete for a single reconstruction. Our results are obtained in seconds, yet are as detailed and even less noisy.}
\label{fig:tech}
\end{figure}

While not specifically designed for 3D reconstruction, our method achieves state-of-the-art performance also for this task. The evaluation setup is the following: given an input image $I$, we draw one random sample from the back view image generator network, and then feed the pair of front/back images to our 3D reconstruction network. For all methods we extract a textured 3D mesh using Marching Cubes and report numerical results in \cref{tab:numerical_eval_image}.
Furthermore, we show qualitative results in \cref{fig:imageto3d}. 
Notably, our method not only performs on par numerically and qualitatively on reconstructed front views, but also generates highly detailed back view texture and geometry.
Finally, we also compare with the optimization-based method TeCH \cite{huang2024tech} in \cref{fig:tech}.
TeCH produces detailed front and back geometry but also also exhibits problems at times, rooted in its 3D pose estimation method. Most importantly, TeCH runs for several hours per instance, while ours computes results in a single feed-forward pass, in only a few seconds.

\begin{table}[t]
    \centering
    \resizebox{0.85\linewidth}{!}{%
    \def\arraystretch{1.1}%
    \setlength{\tabcolsep}{5pt}
    \begin{tabular}{ccc|cccc|l}
        \multicolumn{3}{c|}{3D Metrics} & \multicolumn{4}{c|}{FID $\downarrow$} & \\
        Ch. $\downarrow$ & IoU $\uparrow$ & NC $\uparrow$ & Color. F & Color B.& Normal F. & Normal B. & \\
        \hline
        7.1 & 0.50 & 0.72 & -- & -- & 29.6 & 57.2 &  PIFu
        \cite{saito2019pifu} \\
        
        6.2 & 0.52 & 0.74 & -- & -- & 26.9 & 52.5 &  PIFuHD
        \cite{saito2020pifuhd} \\
        
        7.0 & 0.48 & 0.72 & -- & -- & 36.1 & 49.0 &  ICON
        \cite{xiu2022icon} \\
        
        3.4 & 0.58 & 0.76 & -- & -- & 18.3 & \csecond{36.1} &  ECON
        \cite{xiu2023econ} \\
        
        \cbest{0.84} & \cbest{0.71} & \cbest{0.87} & \cbest{11.0} & \csecond{40.8} & \cbest{11.2} & 38.8 &  PHORHUM*
        \cite{alldieck2022phorhum} \\
        \hline
        \csecond{0.91} & \csecond{0.70} & \csecond{0.86} & \cbest{11.0} & \cbest{20.0} & \csecond{12.8} & \cbest{25.8} &  Ours \\
    \end{tabular}

    }
    \vspace{2mm}
    \caption{Numerical comparisons with single-view 3D reconstructions methods and ablations of our method. We mark the \best{best} and \second{second best} results. All Chamfer metrics are $\times10^{-3}$. Not all methods generate colors. For fair comparisons, we retrained PHORHUM~\cite{alldieck2022phorhum} using the same data as \method. We observe comparable results in terms of 3D metrics, however ours performs better on generating more realistic and diverse back views and back normals. %
    }
    \label{tab:numerical_eval_image}
    \vspace{-7mm}
\end{table}

\subsection{3D Virtual Try On}
\begin{figure}[t]
\vspace{-0.3cm}
    \centering
    \includegraphics[width=.9\textwidth]{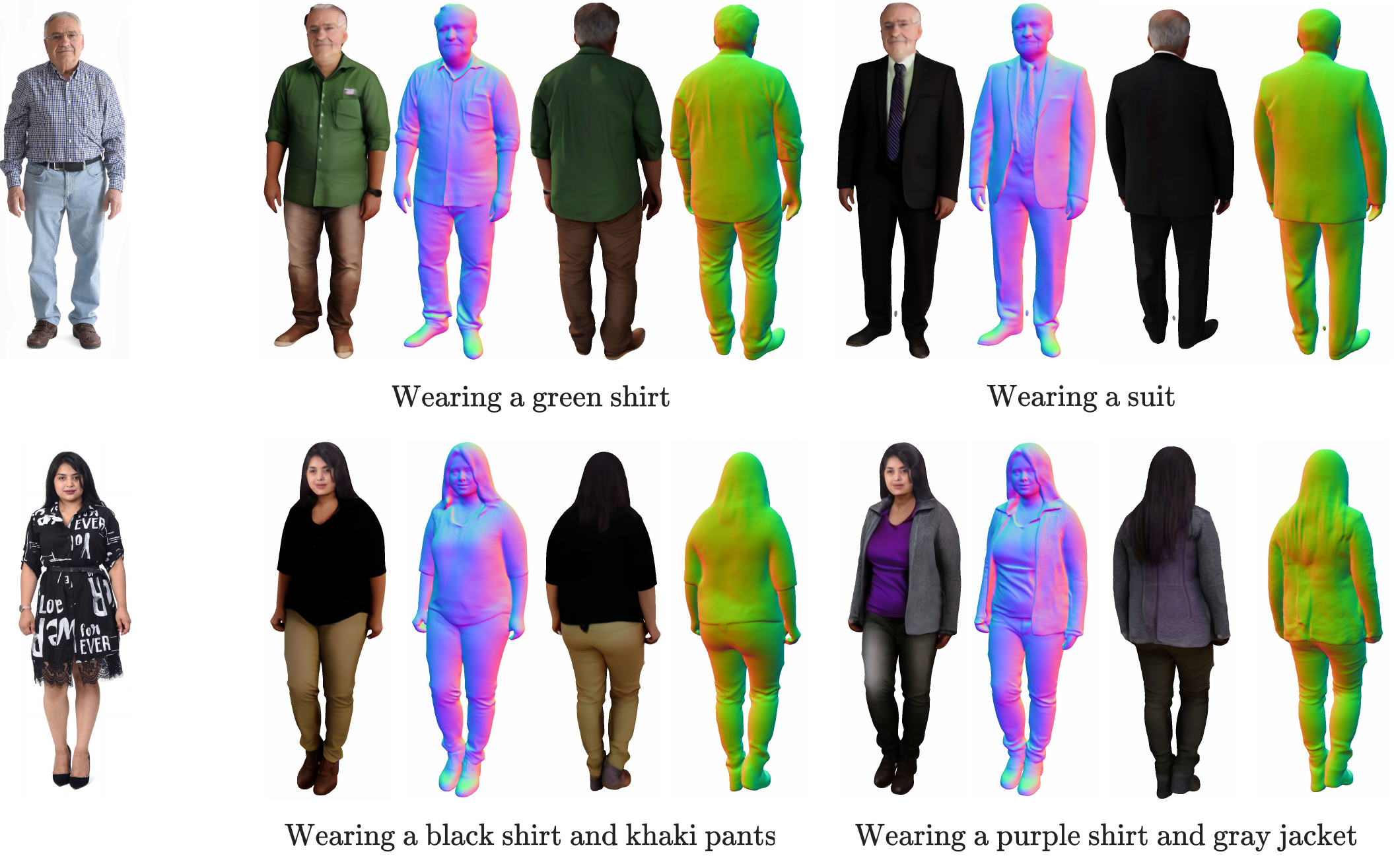} 
    \caption{\textbf{Identity preserving 3D avatar editing.} Our method allows for editing the clothing, while preserving the identity of the generated person. In each row we're using the image on the left as input, recover the person's pose and shape parameters and then run our method using the shown prompts. The generated avatars share the same identity and respect the target prompts.}
\label{fig:editing_examples}
\end{figure}

An immediate application of our method is the option to perform 3D garment edits for a given
identity.
Given an input image, we first recover 3D pose and shape parameters of the person in the image.
Using a text prompt and the identity preservation strategy introduced below, we can generate updated images of the same person wearing different garments or accessories.
To preserve the identity of the person, we first locate the person's head in the source image and use then Repaint \cite{lugmayr2022repaint} to out-paint a novel body for the given head while still conditioning on the estimated pose and shape parameters. With this strategy we ensure that our method not only preserves the facial characteristics but also the overall body proportions.
In \cref{fig:editing_examples} we illustrate such editing examples. The generated 3D edits present garment details like wrinkles on both front and back views, and preserve the subjects' facial appearance. Also, note that body shape and identity are well preserved for the subjects, even though only one image is given. 
While there has been a significant amount of 2D virtual try-on research~\cite{han2018viton, zhu2023tryondiffusion, lee2022high}, our methodology can generate consistent and highly detailed 3D meshes that can be animated and rendered from other viewpoints. %

\subsection{Animation}

\begin{figure}
\vspace{-0.3cm}
    \centering
    \includegraphics[width=.7\textwidth]{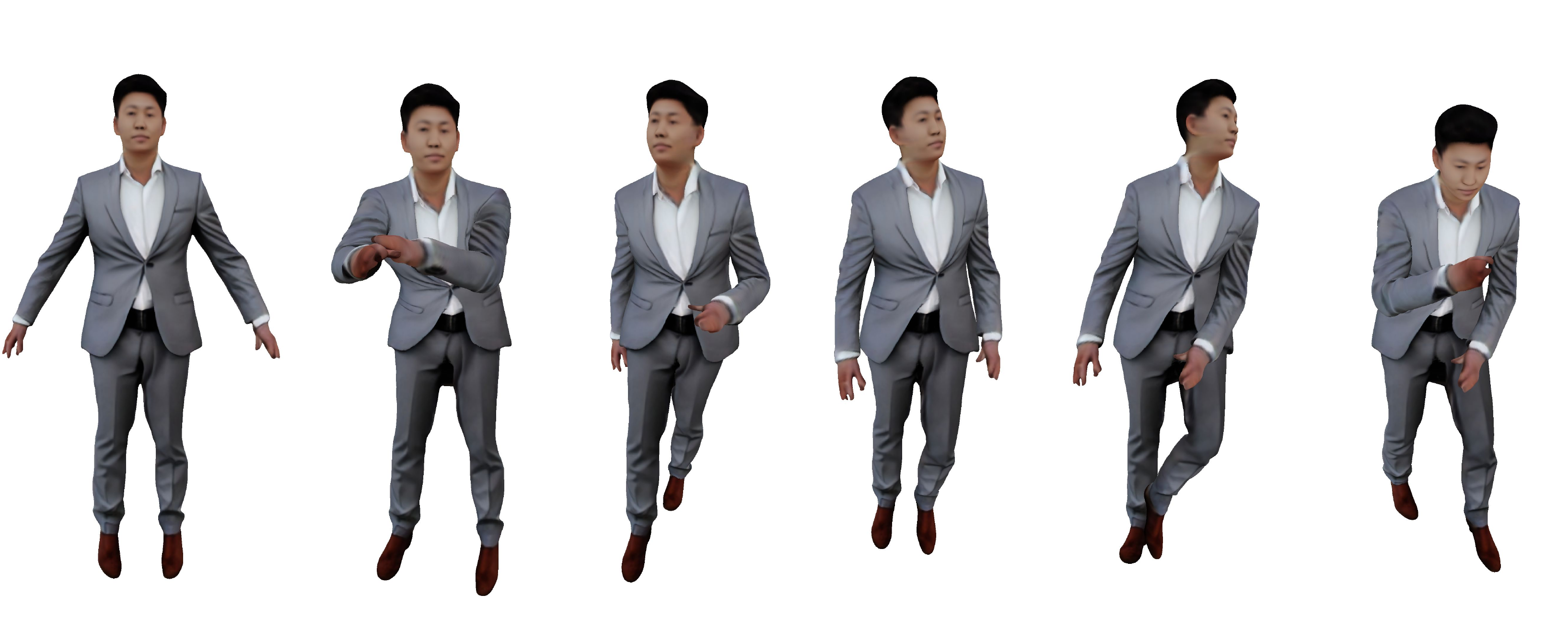} 
    \caption{\textbf{Reposing example.}. We first reconstruct an avatar ``\emph{wearing a gray suit}'' in the A-pose and then we transfer it to different poses.}
\label{fig:animation}
\end{figure}

\method enables animation of the generated assets by design, provided that its generation is conditioned on an underlying body model. In \cref{fig:animation}, we show an example of a generated avatar that is rigged automatically.

\subsection{Ablation Study}

\paragraph{Pose and shape encoding in the 3D reconstruction network.}
We compare the effectiveness of our additional pose and shape encoding inputs $\mathcal{G}$ to our reconstruction network.%
To do so, we use the same set of 100 text prompts as before and for each text prompt sample a random pose and shape configuration. For each $(\tau, \pose, \shape)$ triplet we run inferences and compute 2 meshes: one using only the front and back images, and another one additionally using the dense GHUM encodings. We then evaluate the Chamfer distance between each mesh and the corresponding GHUM mesh. The model using the additional GHUM signals has an average Chamfer distance of $d_{\mathrm{with}} = 1.4$ whereas the one without $d_{\mathrm{without}} = 8.6$, thus validating our design choice. Not only is the control respected well, but this also allows for animation as discussed previously. %

\begin{figure}[ht!]
\vspace{-1mm}
    \centering
    \begin{subfigure}{0.42\linewidth}
    \centering
    \includegraphics[width=0.45\textwidth]
    {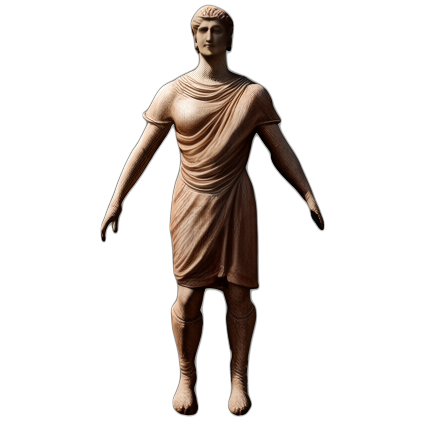}
    \includegraphics[width=0.45\textwidth]
    {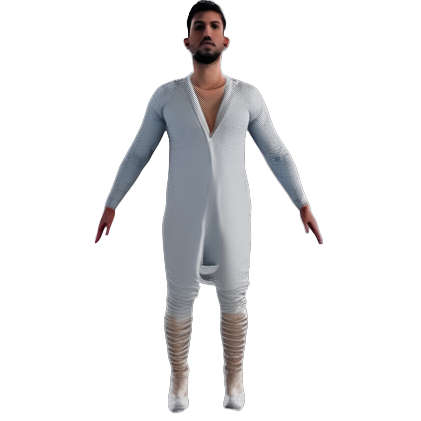}
    \caption*{A Roman marble statue}
    \end{subfigure}
    \begin{subfigure}{0.42\linewidth}
    \centering
    \includegraphics[width=0.45\textwidth]
    {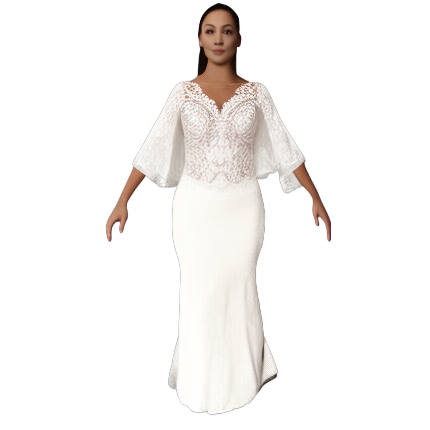}
    \includegraphics[width=0.45\textwidth]
    {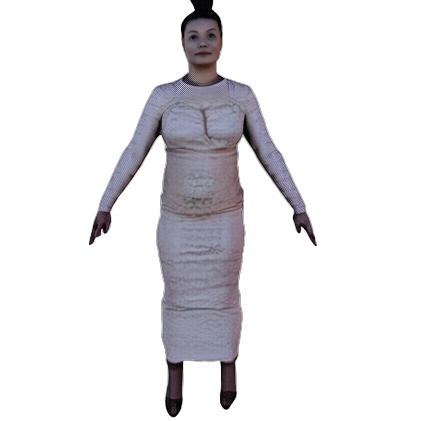}
    \caption*{A person wearing a wedding dress}
    \end{subfigure}
    \vspace{2.5mm}
    \caption{\textbf{Partial \vs complete fine-tuning strategy}. For each text prompt the first example is generated using our partial fine-tuning strategy whereas the second by fine-tuning the entire network. Empirically, partial fine-tuning is more resilient to catastrophic forgetting, generating more diverse and better text-aligned images.}
\label{fig:catastrophic_forgetting}
\vspace{-0.3cm}
\end{figure}

\paragraph{Partial \vs Complete Network Fine-tuning.}
We finetune two Latent Diffusion networks, one in a standard way by optimizing all parameters, and another one using our proposed strategy where we only finetune the convolutional layers of the encoder. Empirically we observed that the network that was finetuned as a whole experienced catastrophic forgetting, and has poor performance when asked to generate types of garment not seen in the training set. \cref{fig:catastrophic_forgetting} shows a comparison for text prompts not in the training set.

%% file: sections/conclusion.tex
\section{Discussion \& Conclusions}

\paragraph{Limitations.}
\method inherits limitations from other pixel-aligned methods, \eg with regions parallel to camera rays being not as detailed in the reconstruction. Further, artifacts might be visible in poses or very loose clothing that are underrepresented during training.

\paragraph{Ethical Considerations.}
We present a generative tool to create 3D human assets, thus reducing the risk of scanning and using real humans for training large scale 3D generative models. \method generates diverse results
and can lead to a better coverage of subject distributions.%

We have presented \method, a novel framework for generating 3D human avatars controlled by text or images and yielding rigged 3D models in 2-10 seconds. 
\method is purely feed-forward, allows for fine-grained control over the generated body pose and shape, and can produce multiple qualitatively different hypotheses. 
\method is composed of a cascade of expert systems, decoupling image generation and 3D lifting. 
Through this design choice, \method benefits from both web-scale image datasets, ensuring high generation diversity, and from  smaller size accurate 3D datasets, resulting in reconstructions with increased detail and precisely controlled based on text and identity specifications.
In the future, we would like to explore other 3D construction strategies besides pixel-aligned features. Longer term, we aim to support highly detailed and controllable 3D human model generation for entertainment, education, architecture and art, or medical applications. 

%% file: sections/supp_content.tex
\section{Implementation Details}

\subsection{Latent Diffusion models}

Our base text-to-image generation model that we finetune is a reimplementation of Stable Diffusion \cite{rombach2022high} with 800 million parameters trained on internal data sources. The latent space has dimensions $64 \times 64 \times 8$ and the input and output images are $512 \times 512 \times 3$. To enable image conditioning in the models, we pass the conditioning image through the latent encoder, and then we concatenate the conditioning latent with the noisy image latent $z_t$ at the input layer. The weights of the input layer are padded with extra channels initialized with zeros to account for the additional 8 input channels from the conditioning.
We train only the parameters of the convolutional layers of the encoder. We finetune the models with the Adam \cite{adam} optimizer using a batch size of 64 images and a learning rate of $5\times 10^{-5}$ on 16 40GB A100 GPUs, for a total of 40000 training iterations. Finetuning takes around 6 hours.
We use both image and text guidance in the style of InstructPix2Pix \cite{brooks2022instructpix2pix}. We use an text guidance weight of 7.5 and an image guidance weight of 2.0. To enable the use of classifier-free guidance, at training time we randomly drop the conditionings. We mask the input text, input image, or input image and text in a mutual exclusive way, with probability $0.05$ each.

\subsection{3D Reconstruction model}
Our network architecture is similar to PHORHUM \cite{alldieck2022phorhum}. The only modification in the encoder the number of input channels, which is $6$ in the case of the front-back model and $12$ for the model with the additional GHUM conditioning $\mathcal{G}$. Additionally, because of the ambiguity of lighting estimation, we drop the shading-albedo decomposition and output the shaded color directly. We train our model for $500$K iterations with the Adam optimizer using a batch size of 32 and a learning rate of $10^{-4}$. Training the model takes 42 hours on 16 40GB A100 GPUs. We use a subset of the original PHORHUM losses for training. We keep the on-surface loss $\mathcal{L}_g$, inside-outside loss $\mathcal{L}_l$, eikonal loss $\mathcal{L}_e$, and color losses $\mathcal{L}_a$ where we replace the albedo $\boldsymbol{a}$ with the shaded color $\boldsymbol{c}$. We omit the rendering losses $\mathcal{L}_r$ as in our setting we did not find them to be useful and also made training slower.

\section{Generation Diversity}

In this section we evaluate the diversity of the generations for the different text-to-3D generation methods. We use the same set of 100 generations used in the main paper. As a proxy for diversity we measure the face similarity of the generated subjects. To quantify the face similarity we use the FaceNet embeddings \cite{schroff2015facenet}. More specifically, we detect and crop the head regions and then use FaceNet to compute the face embeddings. Given images $I_i$ and $I_j$ with embeddings $\mathbf{e}_i$, $\mathbf{e}_j$ respectively, their pairwise similarity is defined as $s_{ij} = \mathbf{e}_i^T \mathbf{e}_j \in [0,1]$. The average pairwise similarity $\mathcal{S}$ over a set of images $\mathcal{D}$ is then defined as:
\begin{equation}
\mathcal{S} = \frac{\sum_{I_i \in \mathcal{D}}\sum_{I_j \in \mathcal{D} \backslash \{I_i\}} s_{ij}}{|\mathcal{D}| \cdot (|\mathcal{D}| -1 )}.
\end{equation}
Intuitively, a high value of $\mathcal{S}$ means that the generated faces are similar to each other.
We also consider the maximum pairwise similarity between an image $I_i$ and a reference dataset $\mathcal{D}$ defined as $s_{i} = \max_{I_j \in \mathcal{D}} s_{ij}$.

The similarity metric $s_{i}$ quantifies whether the face in image $I_i$ is similar to some face from dataset $\mathcal{D}$.

As shown on \tablename~\ref{tab:comp_sim} our method is able to generate more diverse faces than representative optimization-based methods like DreamHuman \cite{kolotouros2023dreamhuman} or TADA\cite{liao2023tada}. We use the similarity between 100 randomly selected training subjects as reference, and our method is able to generate people with comparable similarity scores.

Additionally, we test whether our method overfits on training identities, by comparing the average maximum similarities of our generations with $200$ randomly sampled 3D models from the training and test sets. As reported on \tablename~\ref{tab:train_val_sim}, our generated avatars have the same pairwise similarity scores with either models, thus showing that our model did not overfit on the training set identities.

\begin{table}[t]
    \centering
    \resizebox{0.7\linewidth}{!}{%
        \def\arraystretch{1.1}%
        \setlength{\tabcolsep}{3pt}
        \begin{NiceTabular}{c|l}[colortbl-like]
            Average pairwise face similarity $\mathcal{S}$ $\downarrow$ & \\
            \hline
            \csecond{0.64} & DreamHuman \cite{kolotouros2023dreamhuman}\\
            0.78& TADA \cite{liao2023tada}\\
            \cbest{0.62} & Ours\\ \hline
            0.55 & Random training subjects (lower bound)\\
        \end{NiceTabular}
    }
    \caption{\textbf{Generation diversity evaluation.}  We mark the \best{best} and \second{second best} results. Our method is able to generate faces with larger diversity than the baselines. For reference, we also report the average face similarity for training subjects.}
    \label{tab:comp_sim}
\end{table}

\begin{table}[t]
    \centering
    \resizebox{0.8\linewidth}{!}{%
        \begin{NiceTabular}{cc|l}[colortbl-like]
            Average Maximum Similarity & Median Maximum Similarity & \\
            \hline
            0.67 & 0.67 & Train subjects\\
            0.67 & 0.67 & Test subjects\\ \hline
            0.88 & 0.81 & Generations (self-similarity)\\
        \end{NiceTabular}
    }
    \vspace{1mm}
    \caption{\textbf{Evaluating training set memorization.} We evaluate the similarity of our generated faces with those from the training and test sets. The results show that our model did not overfit on the training set identities.}
    \label{tab:train_val_sim}
\end{table}

\section{Additional Qualitative Results}

In this section we show additional qualitative results that we could not include in the main paper due to space constraints.

\subsection{Relightable Avatar Generation}
\begin{figure}
    \centering
    \includegraphics[width=\textwidth]{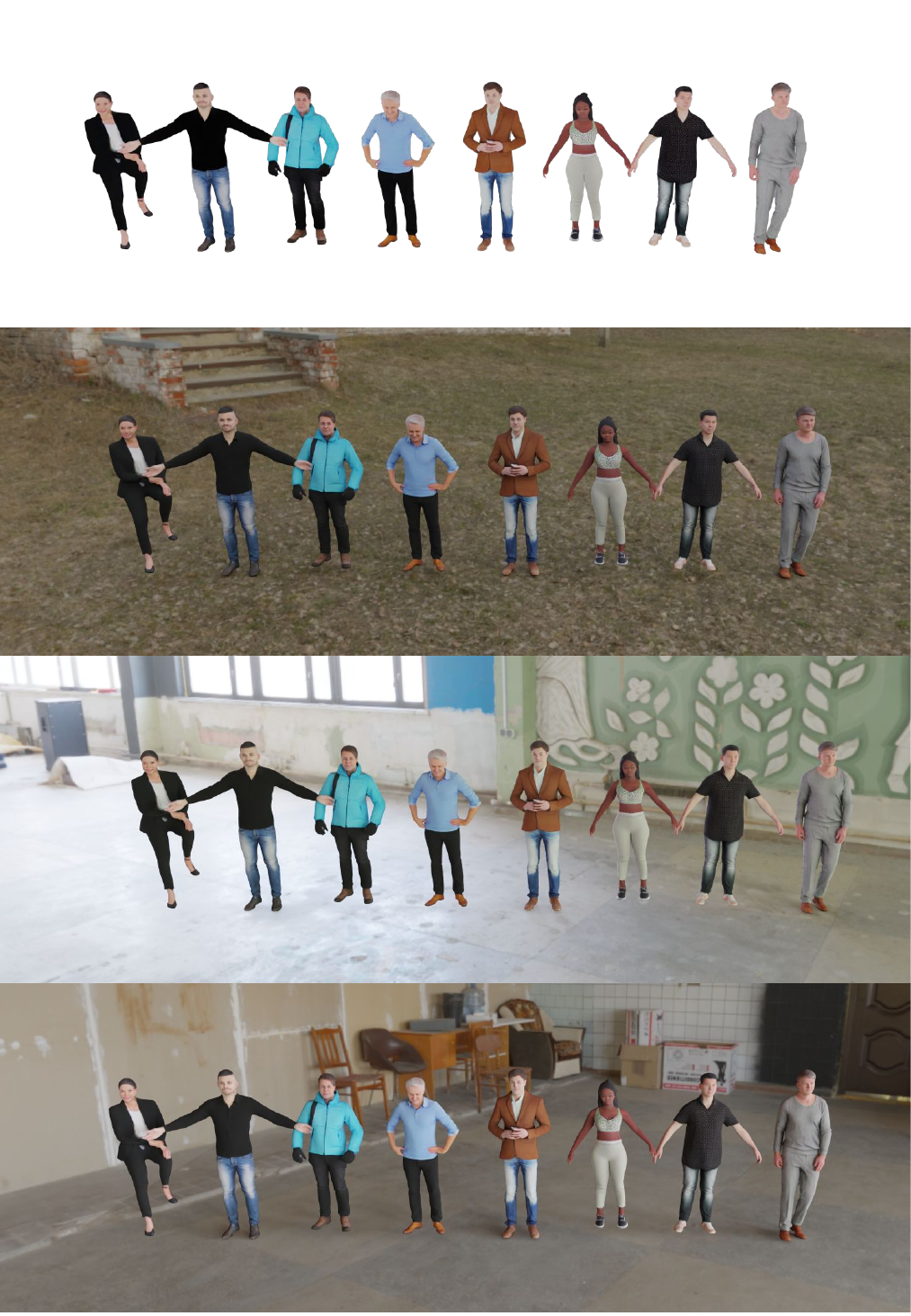} 
    \caption{\textbf{Albedo generation and relighting.} The first row shows 8 avatars generated by probing our method to generate albedo instead of shaded colors. The next 3 rows shows the results of rendering the avatars in different HDRI environments.}
\label{fig:relight}
\end{figure}

By design, our image generation models produce images of people with shading. We additionally experiment with generating albedo images instead of shaded ones, and in this way we can create 3D avatars that can then be relighted in different environments. We teach our model to produce albedo images by randomly substituting the shaded model renderings with unshaded ones at training time, and also appending \emph{``uniform lighting''} to the text prompt. In \figurename~\ref{fig:relight} we show example generations that are rendered in different HDRI environments.

\subsection{Semantic editing}
\begin{figure}[t]
    \centering
    \includegraphics[width=.9\textwidth]{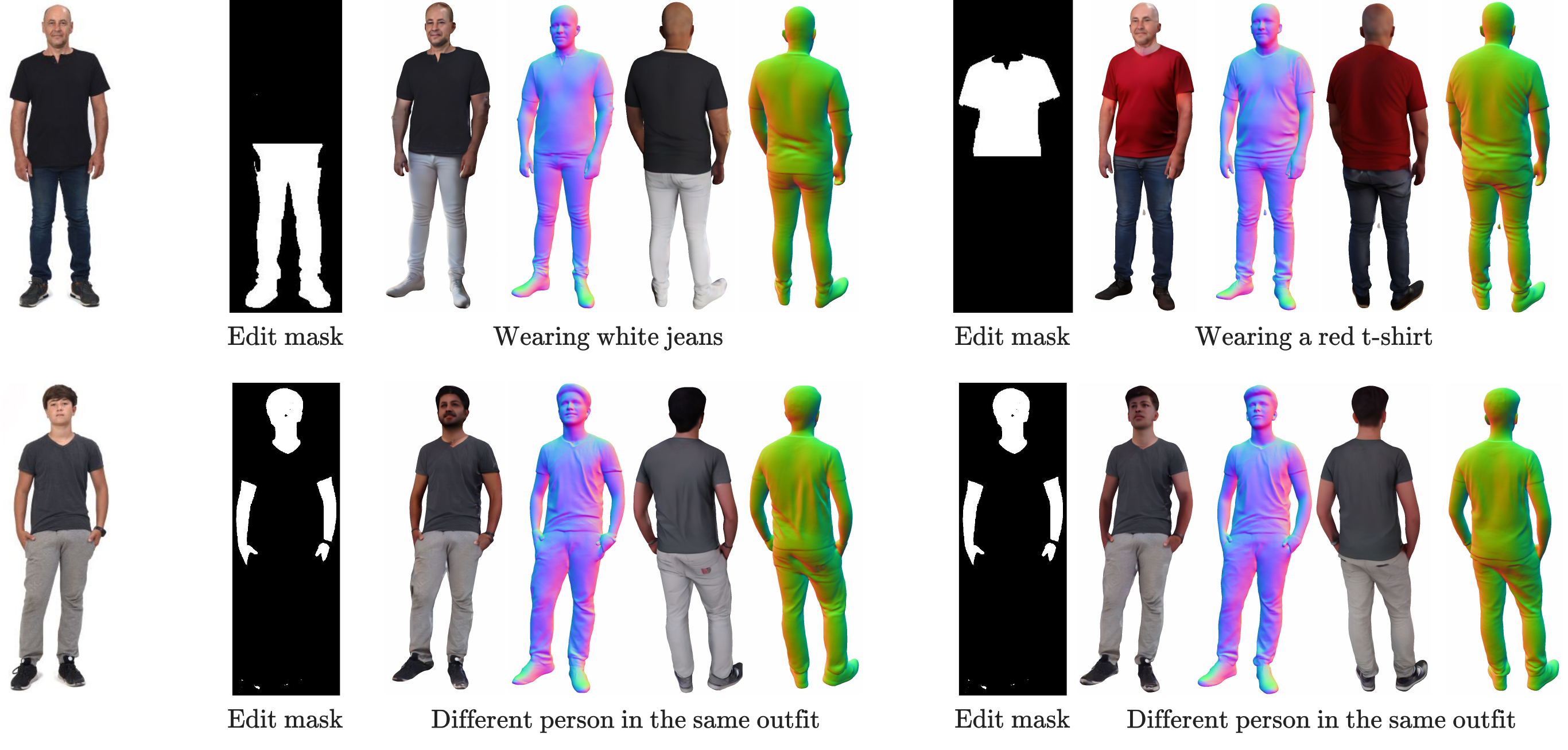} 
    \caption{\textbf{Semantic editing.} In each row we're using the image on the left as input, recover the person's pose and shape parameters and then run our method on updated prompts using the same pose and shape conditioning. In the first row we use additional clothing segmentation masks to perform edits only in specific body regions. In the second row we mask out the person and then generate new people in the same pose wearing the same outfit.}
\label{fig:suppmat_editing_examples}
\end{figure}

We show additional results for the task of semantic editing. We explore 2 scenarios: changing only specific garments on the body while preserving the rest of the appearance and changing the identity of the person wearing the outfit. The input to our method is an image of a person, editing instructions and corresponding semantic segmentation masks. The results are shown on \figurename~\ref{fig:suppmat_editing_examples}

